\documentclass[letterpaper, 10 pt, conference]{ieeeconf}  
\IEEEoverridecommandlockouts                            
\usepackage{graphics} 
\usepackage{epsfig} 
\usepackage{times} 
\usepackage{cite}
\usepackage{amsmath} 
\usepackage{amssymb}  
\usepackage{amsfonts}
\usepackage{algorithm}
\usepackage{algpseudocode}
\usepackage{graphicx}
\usepackage{subcaption}
\usepackage{xcolor}
\usepackage{url}

\xdefinecolor{redZ}{RGB}{234,29,93}
\xdefinecolor{blueZ}{RGB}{19,106,213}
\xdefinecolor{yellowZ}{RGB}{251,138,46}
\xdefinecolor{greenYL}{RGB}{0, 191, 0}

\input{abkuerzung.sty}

\IEEEoverridecommandlockouts                              

\usepackage[utf8]{inputenc}
\usepackage{leftidx}
\usepackage{bm}
\usepackage{amssymb,amsmath,amsfonts,empheq}
\usepackage{graphicx}
\usepackage{siunitx}
\usepackage{multicol}
\usepackage{algorithm}
\usepackage{algpseudocode}
\usepackage{type1cm}
\usepackage{pifont}
\usepackage{lettrine}
\usepackage{color}
\usepackage{cite}
\usepackage{breqn}
\usepackage{tikz}
\usepackage{array}
\usepackage{import}
\usepackage{amsmath}
\usepackage{amssymb}
\usepackage{bm} 
\usepackage{booktabs}
\usepackage{optidef}
\usepackage{graphicx}
\usepackage{xcolor}
\usepackage{siunitx}
\usepackage{fixltx2e}
\usepackage{multirow}
\usepackage{cite}
\usepackage{import}
\usepackage[draft]{hyperref}
\usepackage{microtype}
\usepackage{tabularx}
\usepackage{booktabs}
\usepackage{comment}
 \usepackage{microtype}
 \usepackage{xifthen}

\usepackage{multirow}
\usepackage{scalerel}
\usepackage{fixltx2e}

\usepackage{xfp}
\usepackage{makecell}

\makeatletter
\renewcommand{\section}{\@startsection{section}{1}{\z@}{0.5ex plus 0.5ex minus 0.3ex}%
	{0.4ex plus 0.6ex minus 0ex}{\normalfont\normalsize\centering\scshape}}%
\renewcommand{\subsection}{\@startsection{subsection}{2}{\z@}{0.5ex plus 0.5ex minus 0.4ex}%
	{0.3ex plus 0.5ex minus 0ex}{\normalfont\normalsize\itshape}}%
\makeatother

\newcommand{\tofullversion}[1]{}  

\title{\LARGE \bf
Contact-Safe Reinforcement Learning \\ with ProMP Reparameterization and Energy Awareness
}

\author{
Bingkun Huang, Yuhe Gong, Zewen Yang, Tianyu Ren*, Luis Figueredo 
\thanks{
This work was funded by the Lighthouse Initiative Geriatronics by StMWi
Bayern (Project X, Grant No. 5140951).
B. Huang and T. Ren are with  Munich Institute of Robotics \& Machine Intelligence, Technische Universität München (TUM), Germany. 
Y. Gong and L. Figueredo are with the School of Computer Science at The University of Nottingham. L. Figueredo is also an Associated Fellow at the MIRMI-TUM. \newline
*Corresponding author: T. Ren \textless{}{\tt\small tianyu@robot-learning.de}\textgreater{}
}
}

\begin{document}

\maketitle
\thispagestyle{empty}
\pagestyle{empty}


\author{
Anonymous submission
}


\maketitle
\begin{abstract}
Reinforcement learning (RL) approaches based on Markov Decision Processes (MDPs) are predominantly applied in the robot joint space, often relying on limited task-specific information and partial awareness of the 3D environment. 
In contrast, episodic RL has demonstrated advantages over traditional MDP-based methods in terms of trajectory consistency, task awareness, and overall performance in complex robotic tasks. 
Moreover, traditional step-wise and episodic RL methods often neglect the contact-rich information inherent in task-space manipulation, especially considering the contact-safety and robustness. 
In this work, contact-rich manipulation tasks are tackled using a task-space, energy-safe framework, where reliable and safe task-space trajectories are generated through the combination of Proximal Policy Optimization (PPO) and movement primitives. 
Furthermore, an energy-aware Cartesian Impedance Controller objective is incorporated within the proposed framework to ensure safe interactions between the robot and the environment.
Our experimental results demonstrate that the proposed framework outperforms existing methods in handling tasks on various types of surfaces in 3D environments, achieving high success rates as well as smooth trajectories and energy-safe interactions.

\end{abstract}

\IEEEpeerreviewmaketitle

\section{Introduction}

Contact-rich robotic manipulation imposes stringent requirements on safety, adaptability, and robustness due to discontinuous dynamics, transient contact forces, and complex energy exchanges. Uncontrolled interaction can lead to instability, excessive forces, or unintended motion, posing risks to both the robot and its environment. Ensuring safe physical interaction therefore requires not only effective regulation of energy flow, but also the ability to generate smooth and adaptable trajectories that remain robust under uncertainty. Traditional model-based motion-generation approaches (e.g., Movement Primitives) rely on accurate models and measurements \cite{ArmstrongHelouvry1994,Jean1999}, yet such accuracy is difficult to guarantee in practice, particularly for physical interaction tasks. Reinforcement learning (RL), on the other hand, offers robustness through data-driven exploration and training (e.g., via domain randomization), but often produces non-smooth stepwise policies and lacks explicit safety guarantees.

\begin{figure}[t]
    \centering
    \includegraphics[width=1\linewidth]{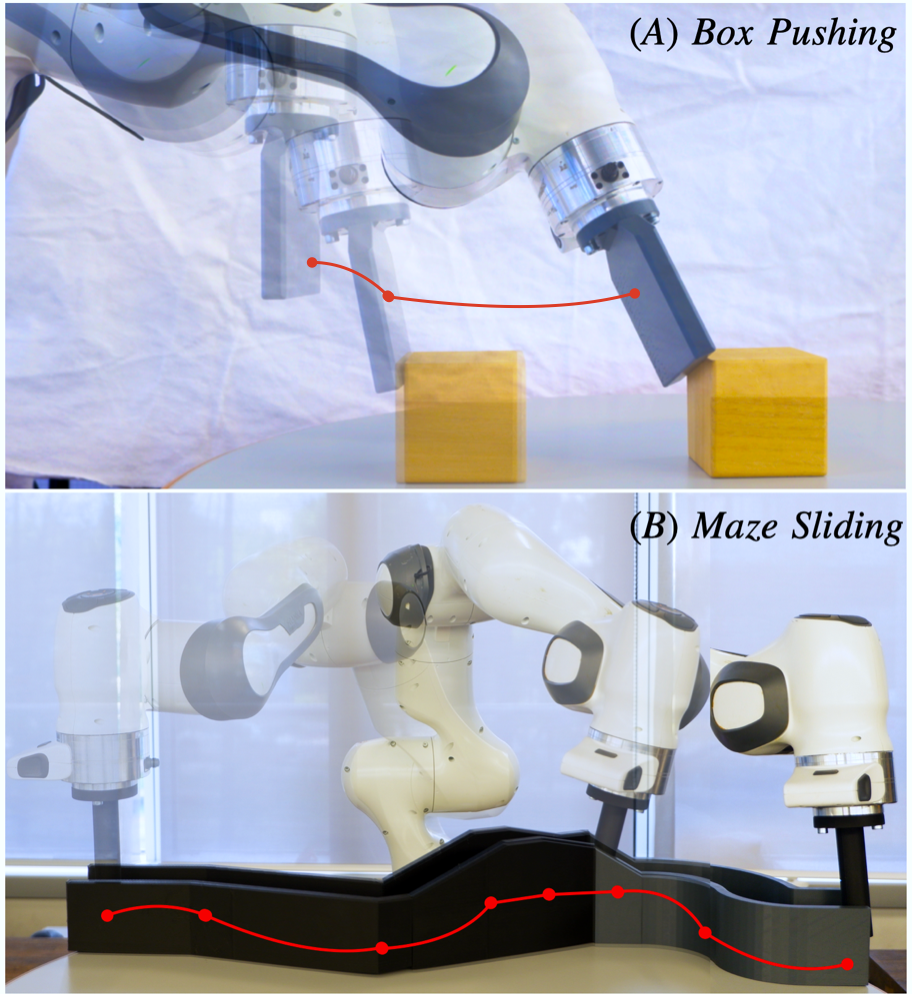}
    \caption{Contact-rich manipulation tasks studied in this paper: (A) box pushing with sustained surface contact; (B) maze sliding where the tool navigates unseen turns relying on contact feedback. These tasks require regulating interaction energy to avoid unsafe force or power bursts during exploration and execution.}
    \label{fig:intro}
\end{figure}

To address these challenges, several key capabilities are required for a safe smooth manipulation framework: (1) contact-aware representations that capture task-space constraints and uncertainties; (2) trajectory-level planning strategies that ensure smooth, dynamically feasible motions, such as movement primitives; and (3) explicit mechanisms to regulate energy exchange, ensuring interactions remain within safe operating bounds through passivity-based and energy-aware control. Separately, each of these is well studied, yet no existing framework tightly integrates data-driven robustness, trajectory-level smoothness, and passivity-based safety for contact-rich manipulation. This work bridges that gap by embedding RL into a task-space, movement-primitive representation while explicitly enforcing energy-safe interaction during both learning and execution. We study two representative contact-rich tasks—box pushing and maze sliding—as illustrated in Fig.~\ref{fig:intro}.

\subsection{Related Work}
Despite the growing success of RL for manipulation~\cite{Lillicrap2015DDPG,Schulman2017PPO,Levine2016EndToEnd}, its application to contact-rich tasks remains relatively limited~\cite{Kober2013Survey}. Most RL methods are developed and evaluated in joint space or quasi-contact scenarios, where physical interactions with the environment are minimal and contact forces can often be neglected. Contact-rich manipulation (e.g., pushing, sliding, or assembly) introduces discontinuous dynamics and complex energy exchanges, which pose significant challenges for standard RL algorithms. Existing work has explored either model-based approaches that incorporate simplified contact dynamics or force feedback into policy learning~\cite{Kormushev2013RLRoboticsSurvey,Pastor2013MotorPrimitives}, or episodic RL combined with movement primitives to improve trajectory consistency under contact constraints~\cite{Rajeswaran2017DAPG,Otto2023BBRLMP}. However, these approaches are still scarce and often struggle with generalization to unseen contacts and with explicit energy management on real hardware.

Safe reinforcement learning (SafeRL)~\cite{Garcia2015SafeRL} aims to enforce safety constraints during both learning and execution, such as limits on peak force, power, or energy. In contact-rich manipulation, transient forces, frictional uncertainties, and discontinuous dynamics make precise constraint modeling difficult, often limiting SafeRL's effectiveness. Complementarily, passivity-based control and energy-tank frameworks provide a principled safety layer by preventing the robot from injecting uncontrolled energy during physical interactions~\cite{Ortega2002,Haddadin2017,DeLuca2019}. Yet, purely passivity-based approaches can be conservative and may not optimize task performance or trajectory smoothness.

Recent work has addressed complementary aspects of this problem. To improve trajectory consistency and handle sparse rewards, Otto et al.\ showed that policies parameterizing movement primitives can generate smoother trajectories than low-level stepwise policies~\cite{pmlr-v205-otto23a}. For better generalization to unforeseen geometries, Zhou et al.\ proposed via-point movement primitives that adapt trajectories to novel constraints~\cite{Zhou2019ViaPointMP}. From a safety standpoint, passivity/energy-based methods explicitly regulate interaction energy; for example, Zhang et al.\ introduced energy constraints to guarantee safe interactions, though outside a trajectory-level policy-learning context~\cite{Zhang2025PassivityCentric}. Despite these advances, there remains no framework that jointly couples RL with trajectory-level movement primitives and energy-/passivity-aware execution for contact-rich manipulation.

\subsection{Contribution}
We propose \textbf{PPT}, a contact-safe RL framework that integrates Probabilistic Movement Primitives (ProMPs) for smooth task-space trajectory generation, Proximal Policy Optimization (PPO) for adaptive modulation, and an energy-tank passivity layer for energy-safe interaction in contact-rich manipulation. Our key contributions are:
\textbf{(C1)} A task-space RL formulation that parameterizes actions in a low-dimensional ProMP weight space and executes them through Cartesian impedance control, enabling smooth, compliant trajectories for contact-rich tasks.
\textbf{(C2)} A real-time energy-aware passivity controller (energy tank) that constrains interaction power/energy, providing safety guarantees during both learning and deployment under discontinuous contact dynamics.
We validate the approach in simulation and on a Franka Panda robot on box-pushing and maze-sliding tasks, with ablations demonstrating the role of each component in safety, smoothness, and task success.

\section{Preliminary}
\label{sec:background}

Robotic manipulation in contact-rich environments requires reasoning about trajectories, control forces, and safety simultaneously. 
In this section, we introduce the core concepts underlying our approach: trajectory representation, reinforcement learning for control adaptation, physical safety via passivity, and impedance-based execution.

\subsection{Trajectory Representation with ProMPs}

To model complex trajectories in a compact and adaptable way, we employ \emph{Probabilistic Movement Primitives (ProMPs)}~\cite{Paraschos2013ProMP}. 
ProMPs encode a distribution over trajectories rather than a single deterministic path, allowing the robot to capture variability observed in demonstrations. 

Let $\phi \in [0,1]$ denote a canonical phase variable, and $\boldsymbol{\Phi}(\phi) \in \mathbb{R}^{D \times K}$ a set of basis functions (e.g., radial basis functions). 
A trajectory $\boldsymbol{y}(\phi) \in \mathbb{R}^D$ is expressed as
\begin{equation}
\label{eq::promp}
\boldsymbol{y}(\phi) = \boldsymbol{\Phi}(\phi) \boldsymbol{w}, \quad 
\boldsymbol{w} \sim \mathcal{N}(\boldsymbol{\mu}_w, \boldsymbol{\Sigma}_w),
\end{equation}
where $\boldsymbol{w} \in \mathbb{R}^K$ are the trajectory weights. 
The Gaussian prior over $\boldsymbol{w}$ captures demonstration variability, 
while the linear combination ensures smoothness and computational tractability. 
In our method, we extend ProMPs with via-point conditioning and reinforcement-learning-based residual updates (Sec.~\ref{sec:method}).

\subsection{Reinforcement Learning with PPO}

While ProMPs provide structured priors, adapting them to new environments or tasks requires learning-based refinement. 
We model the control problem as a Markov Decision Process (MDP) 
$\mathcal{M}=(\mathcal{S},\mathcal{A},P,r,\gamma)$, where 
$\mathcal{S}$ is the state space, $\mathcal{A}$ the action space, $P$ the transition dynamics, 
$r$ the reward function, and $\gamma$ the discount factor. 

A stochastic policy $\pi_\theta(a|s)$ outputs a probability distribution over actions given a state. In the remainder, we denote the executed robot command as $u_t$; in generic RL notation it corresponds to the action $a_t$.
We optimize the policy using \emph{Proximal Policy Optimization (PPO)}~\cite{Schulman2017PPO}, 
which stabilizes learning by limiting the magnitude of policy updates. 
The clipped surrogate objective is
\begin{align}
\mathcal{L}_{\text{clip}}(\theta)
&= \mathbb{E}_t\Big[\min\big(r_t \hat{A}_t, \bar{r}_t \hat{A}_t\big)\Big], \\
\bar{r}_t &\coloneqq \operatorname{clip}(r_t, 1-\epsilon, 1+\epsilon), \quad
r_t \coloneqq \frac{\pi_\theta(a_t|s_t)}{\pi_{\theta_{\text{old}}}(a_t|s_t)},
\end{align}
where $s_t \in \mathcal{S}$, $a_t \in \mathcal{A}$, 
$\hat{A}_t$ is the estimated advantage (e.g., via GAE), and $\epsilon$ is a small clipping parameter. 
PPO refines ProMP-generated references to improve task performance while maintaining stable learning.

\subsection{Safety via Passivity and Energy-Tank Mechanisms}

When interacting with the environment or humans, safety is critical. 
We enforce \emph{passivity}, which ensures that the robot cannot inject unbounded energy~\cite{Hogan1985PassiveSystems,Ortega2001Passivity}. 
Let $E \in \mathbb{R}_{\ge 0}$ denote the robot's stored energy (kinetic + potential), and $p \in \mathbb{R}$ the instantaneous power exchanged with the environment. 
Passivity is expressed as
\begin{equation}
\dot{E} \leq p.
\end{equation}
This constraint prevents high-impact forces, actuator saturation, and instability. 
Energy-tank mechanisms~\cite{Haddadin2012EnergyTanks} explicitly enforce this bound, ensuring safe execution of learned or planned trajectories.

\subsection{Cartesian Impedance Control}

Finally, to execute trajectories while interacting with the environment, we employ Cartesian impedance control. 
Let $(\boldsymbol{x}_d, \boldsymbol{R}_d)$ be the desired end-effector pose, 
with position error $\boldsymbol{e}_x = \boldsymbol{x}_d - \boldsymbol{x}$ and orientation error 
$\boldsymbol{e}_R = \frac{1}{2} (\boldsymbol{R}_d^\top \boldsymbol{R} - \boldsymbol{R}^\top \boldsymbol{R}_d)^\vee$, 
where $\boldsymbol{x} \in \mathbb{R}^3$, $\boldsymbol{R} \in SO(3)$, and $(\cdot)^\vee$ maps a skew-symmetric matrix to a vector. Let $\boldsymbol{\omega}$ and $\boldsymbol{\omega}_d$ denote the current and desired end-effector angular velocities, and define the angular-velocity error $\boldsymbol{\omega}_e := \boldsymbol{\omega}_d-\boldsymbol{\omega}$.

A standard Cartesian impedance law is
\begin{equation}
\boldsymbol{f}=\boldsymbol{K}_p\,\boldsymbol{e}_x+\boldsymbol{K}_d\,\dot{\boldsymbol{e}}_x,\qquad
\boldsymbol{\tau}=\boldsymbol{K}_{p,R}\,\boldsymbol{e}_R+\boldsymbol{K}_{d,R}\,\boldsymbol{\omega}_e .
\end{equation}
Here $\boldsymbol{f}\in\mathbb{R}^3$ and $\boldsymbol{\tau}\in\mathbb{R}^3$ are the commanded Cartesian
force and torque at the end-effector. The resulting wrench and executed joint torques are
\[
\boldsymbol{\lambda}=\begin{bmatrix}\boldsymbol{f}\\ \boldsymbol{\tau}\end{bmatrix}\in\mathbb{R}^6,
\qquad
\boldsymbol{\tau}_{\text{imp}}=\boldsymbol{J}^\top(\boldsymbol{q}_t)\,\boldsymbol{\lambda},
\]
where $\boldsymbol{J}(\boldsymbol{q}_t)$ is the $6{\times}n$ geometric Jacobian at joint configuration $\boldsymbol{q}_t$, and $\boldsymbol{K}_p,\boldsymbol{K}_d,\boldsymbol{K}_{p,R},\boldsymbol{K}_{d,R}$ are positive-definite gains.
This controller executes ProMP- and PPO-generated references while maintaining compliance and safety.

\begin{figure*}[t]
    \centering
    \includegraphics[width=1\linewidth]{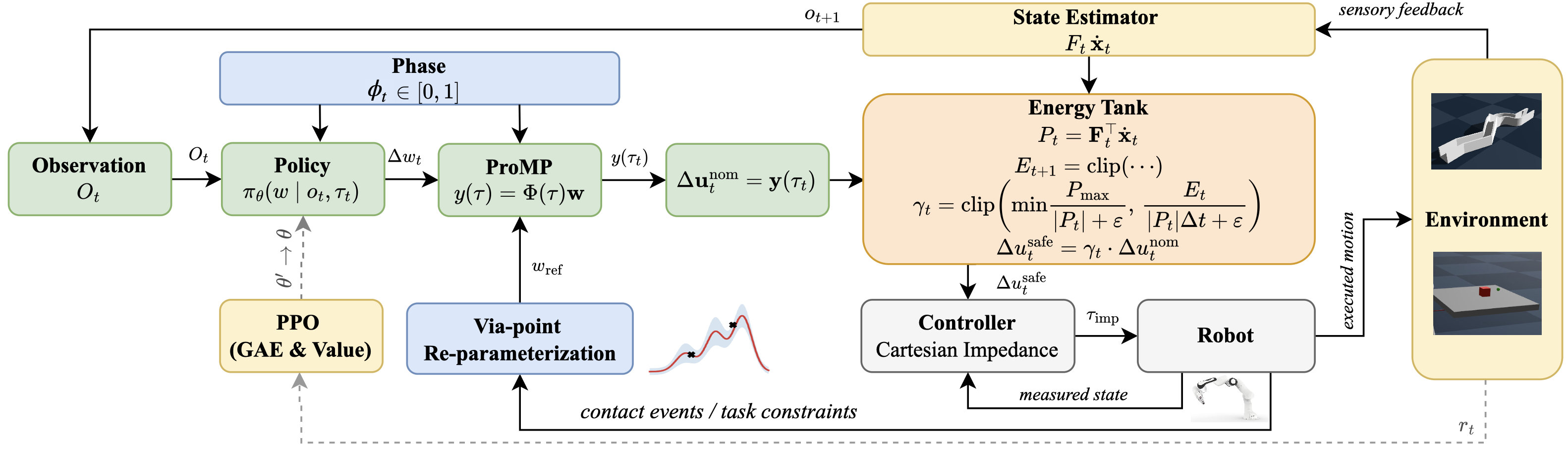}
    \caption{Overview of the PPT framework. Observations and a phase variable are fed to the policy, which outputs residual ProMP weights. These are combined with via-point conditioned reference weights to form a ProMP trajectory. An energy-tank layer scales the command to ensure safe execution, which a Cartesian impedance controller tracks and converts to joint torques. Interaction feedback informs via-point reparameterization. PPO updates the policy using GAE and a value critic (dashed arrow).}
    \label{fig:pipeline}
\end{figure*}

\section{Problem Definition}
\label{sec:problem}

\subsection{Overview of the System}

Our framework, \textit{PPT} (\textbf{P}roMP \textbf{P}PO Energy-\textbf{T}ank), integrates three complementary components: 
(i) \textbf{ProMPs} for structured, low-dimensional trajectory representation; 
(ii) \textbf{PPO} for adaptive, learning-based trajectory refinement; 
and (iii) \textbf{Energy-tank passivity control} for safe execution. 
An overview of the PPT framework is shown in Fig.~\ref{fig:pipeline}.
The system operates as follows:
\begin{enumerate}
    \item \textbf{Trajectory Representation:} ProMPs encode robot trajectories as distributions over basis functions, allowing smooth and low-dimensional motion representation.
    \item \textbf{Policy Refinement:} PPO updates the ProMP weights at each timestep,
\begin{align}
    \Delta \boldsymbol{w}_t = \pi(o_t),
\end{align}
    to adapt trajectories based on observed performance.
    \item \textbf{Safe Execution:} Trajectories generated from ProMP weights,
\begin{align}
        \boldsymbol{y}_t = \boldsymbol{\Phi}(\phi_t) \boldsymbol{w}_t,
\end{align}
    are updated online according to energy feedback. The energy-tank mechanism ensures interaction forces remain within safe limits.
    \item \textbf{Online Adaptation:} At test time, partial task-specific constraints can be provided via a small set of via-points
\begin{align}
    D_t = \{(\phi_i, y_i, \Sigma_i)\}_{i=1}^{N_t},
\end{align}
    which are incorporated by a conditioning operator $\mathcal{C}$:
\begin{align}
    \boldsymbol{w}_t^\star = \mathcal{C}(\boldsymbol{w}_t; D_t), \quad \boldsymbol{y}_t = \Phi(\phi_t) \boldsymbol{w}_t^\star.
\end{align}
\end{enumerate}

This integration allows the system to generate \emph{smooth, adaptive trajectories} while maintaining \emph{safe energy interactions} and \emph{generalization to unseen environments}.

\subsection{Problem Statement.}

We consider the problem of learning a policy $\pi_\theta(u_t \mid s_t)$ for a robot in a contact-rich environment, where $s_t \in \mathcal{S}$ denotes the robot state and $u_t \in \mathcal{U}$ the control input at time $t$. The robot interacts with dynamic and partially unknown objects, surfaces, or humans. 

The goal is to maximize a cumulative task reward
\[
R = \sum_{t=0}^T r(s_t, u_t),
\]
while ensuring safety and adaptability. The reward encodes four key aspects:
\begin{enumerate}
    \item \textbf{Goal / Path Shaping:} encourages progress toward the task goal and proper path following.
    \item \textbf{Task Generalization:} encourages trajectory adaptation in free space.
    \item \textbf{Contact-phase / Haptic Locomotion:} ensures safe and effective motion during contact.
    \item \textbf{Power / Energy Safety:} penalizes excessive forces or energy usage.
\end{enumerate}

Formally, the problem can be expressed as
\begin{equation}
\begin{aligned}
\label{eq::problemdefinition}
\max_{\pi_\theta}\quad & \mathbb{E}\!\left[\sum_{t=0}^{T} r(s_t,u_t)\right] \\[2pt]
\text{s.t.}\quad 
& s_{t+1}=f(s_t,u_t), \ \  s_0 {\sim}\mathcal{S}_0,\ \ s_T{\in}\mathcal{S}_{\text{goal}}, \  u_t {\in} \mathcal{U}, \\[2pt]
& \underbrace{P_t={\boldsymbol{\lambda}^{\text{nom}}_t}^\top \boldsymbol{\nu}_t,\quad p_t:=|P_t|}_{\text{instantaneous power (6D)}},\\[2pt]
& 0\le E_t \le E_{\max},\quad \gamma_t\in[0,1],\\[2pt]
& \boxed{\;
\gamma_t \le \frac{P_{\max}}{\max(\varepsilon,\,p_t)},\qquad
\gamma_t \le \frac{E_t}{\Delta t\,\max(\varepsilon,\,p_t)}
\;}\\[2pt]
& u_t = \gamma_t\,u_t^{\text{nom}},\qquad u_t^{\text{nom}}=\pi_\theta(o_t),\\[2pt]
& E_{t+1}=\min\{E_{\max},\ \max\{0,\,E_t - \gamma_t\,p_t\,\Delta t\}\}.
\end{aligned}
\end{equation}

where
$\pi_\theta(u_t\mid s_t)$ is the stochastic policy (parameters $\theta$),
$s_t\in\mathcal S$ the robot state,
$u_t\in\mathcal U$ the executed control,
$o_t$ the observation,
$T$ the horizon,
and $r(s_t,u_t)$ the step reward; the expectation is over rollouts induced by $\pi_\theta$ and dynamics $f$.
$f:\mathcal S\times\mathcal U\!\to\!\mathcal S$ is the system dynamics,
$\mathcal S_0$ the initial-state distribution,
$\mathcal S_{\text{goal}}$ the goal set,
and $\mathcal U$ the admissible control set.
$\boldsymbol{\lambda}^{\text{nom}}_t=[\boldsymbol{f}^{\text{nom}}_t;\boldsymbol{\tau}^{\text{nom}}_t]$ is the nominal 6D wrench,
$\boldsymbol{\nu}_t=[\dot{\boldsymbol{x}}_t;\boldsymbol{\omega}_t]$ the 6D twist,
$P_t={\boldsymbol{\lambda}^{\text{nom}}_t}^\top\boldsymbol{\nu}_t$ the instantaneous power,
$p_t:=|P_t|$ its magnitude,
$\Delta t$ the control step,
$\varepsilon>0$ a small constant,
$E_t$ the energy-tank level (capacity $E_{\max}$),
$P_{\max}$ the power limit,
$\gamma_t\in[0,1]$ the safety scaling factor,
$u_t^{\text{nom}}=\pi_\theta(o_t)$ the nominal command,
and $u_t=\gamma_t\,u_t^{\text{nom}}$ the safe command.
(3D special case: replace $\boldsymbol{\lambda},\boldsymbol{\nu}$ with $\boldsymbol{f},\dot{\boldsymbol{x}}$.)

The objective is to learn a policy that can \emph{generalize to unseen geometries}, produce \emph{smooth, compliant trajectories}, and maintain \emph{energy-safe interactions} in contact-rich environments.

\section{PPT: contact-safe RL with ProMP}

\label{sec:method}
Building upon the preliminaries, our method introduces a task-space reinforcement learning framework that enables \emph{online trajectory adaptation} while explicitly enforcing \emph{energy-safe interactions} in contact-rich tasks. The framework integrates three key components: structured trajectory priors, policy-driven adaptation, and passivity-based safety.

\subsection{Trajectory Prior with ProMP}
\label{sec:promp_prior}

We represent a $d$-DoF task-space trajectory using a canonical phase variable $\phi\in[0,1]$ and $K$ radial basis functions (RBFs):
\begin{equation}
y(\phi)=\boldsymbol{\Phi}(\phi)^\top \mathbf{W},\qquad
\Phi_k(\phi)=\exp\!\Big(-\frac{(\phi-c_k)^2}{2\sigma_k^2}\Big),
\end{equation}
where $\mathbf{W}\in\mathbb{R}^{K\times d}$ is the basis-weight matrix (one column per task-space DoF), $c_k$ and $\sigma_k$ denote the centers and widths of RBF, respectively.
We place a Gaussian prior over the vectorized weights $\boldsymbol{w}:=\mathrm{vec}(\mathbf{W})$:
\begin{equation}
\boldsymbol{w} \sim \mathcal{N}(\boldsymbol{\mu}_w,\boldsymbol{\Sigma}_w),
\end{equation}
with $(\boldsymbol{\mu}_w,\boldsymbol{\Sigma}_w)$ estimated from demonstrations or prior rollouts. This provides a smooth, low-dimensional, and probabilistic trajectory model that captures typical motion patterns while allowing variability.

\subsection{Reinforcement Learning in ProMP Weight Space}
\label{sec:ppo_promp}

Instead of acting directly in the raw control space, the policy refines the ProMP by outputting residual updates in weight space:
\begin{equation}
\boldsymbol{w}_t = \boldsymbol{w}_{\text{ref}} + \Delta \boldsymbol{w}_t, \qquad
\Delta \boldsymbol{w}_t \sim \pi_\theta(\cdot \mid \tilde{o}_t),
\end{equation}
where $o_t$ denotes the robot observation and we augment it with the phase variable via $\tilde{o}_t := [o_t,\phi_t]$.
The reference weights $\boldsymbol{w}_{\text{ref}}$ are given by the ProMP prior mean or the via-point posterior (Sec.~\ref{sec:via_point}).
Let $\mathbf{W}_t := \mathrm{unvec}(\boldsymbol{w}_t)\in\mathbb{R}^{K\times d}$. The adapted weights decode to references
\[
\hat{y}_t = \boldsymbol{\Phi}(\phi_t)^\top \mathbf{W}_t
\]
which are mapped to Cartesian impedance commands, yielding the nominal control $u_t^{\text{nom}}$.  
The policy $\pi_\theta$ is trained using PPO with the clipped surrogate objective:
\begin{equation}
\mathcal{L}(\theta) = \mathbb{E}_t \Big[ \min \big( r_t(\theta) \hat{A}_t, \ \mathrm{clip}(r_t(\theta),1-\epsilon,1+\epsilon)\hat{A}_t \big) \Big],
\end{equation}
where $r_t(\theta)$ is the likelihood ratio and $\hat{A}_t$ is the advantage estimate.  
Operating in weight space leverages the structure and smoothness of ProMPs while enabling online trajectory adaptation.

\subsection{Trajectory Posterior via Via-Point Conditioning}
\label{sec:via_point}

\begin{figure}
    \centering
    \includegraphics[width=1\linewidth]{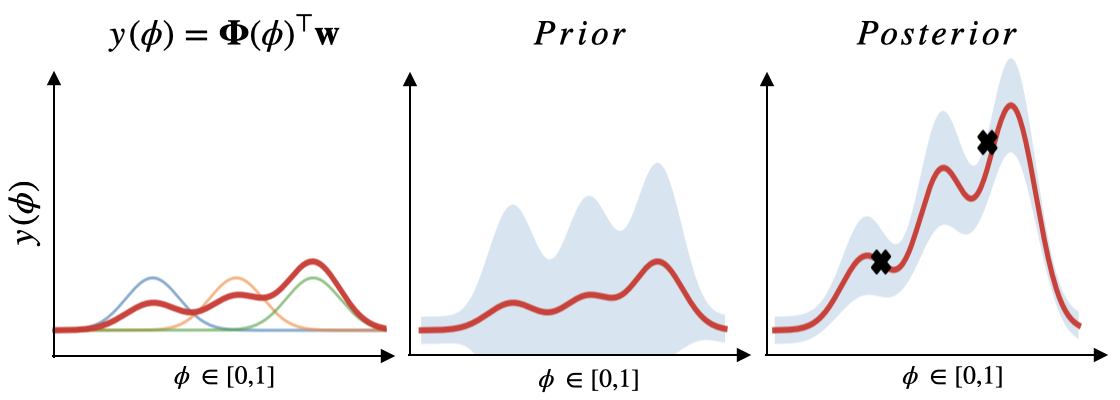}
    \caption{\textbf{ProMP prior and via-point posterior (1D view).}
      Left: trajectory represented as $y(\phi)=\boldsymbol{\Phi}(\phi)^\top \mathbf{w}$ with $K$ RBFs.
      Middle: prior $\mathrm{vec}(\mathbf{w})\sim\mathcal{N}(\boldsymbol{\mu}_w,\boldsymbol{\Sigma}_w)$: mean (red) $\pm 2\sigma$ band (blue).
      Right: posterior conditioned on via-points $\mathcal{D}$ (black $\times$), showing tightened uncertainty near constraints while preserving smoothness.}
    \label{fig:promp_prior_posterior}
\end{figure}

Fig.~\ref{fig:promp_prior_posterior} illustrates the ProMP prior and the via-point posterior update.
To incorporate partial geometric or contact constraints, we condition the prior on a set of via-points $\mathcal{D} = \{(\phi_j, \mathbf{y}_j)\}_{j=1}^M$ with observation covariance $\boldsymbol{\Sigma}_D$. Define
\[
\mathbf{H} = \begin{bmatrix} \boldsymbol{\Phi}(\phi_1)^\top \\ \vdots \\ \boldsymbol{\Phi}(\phi_M)^\top \end{bmatrix},\quad
\mathbf{y}_{\mathcal{D}} = \begin{bmatrix} \mathbf{y}_1 \\ \vdots \\ \mathbf{y}_M \end{bmatrix}.
\]
The posterior distribution over weights is
\begin{align}
\boldsymbol{\Sigma}_{w|\mathcal{D}} &= (\boldsymbol{\Sigma}_w^{-1} + \mathbf{H}^\top \boldsymbol{\Sigma}_D^{-1} \mathbf{H})^{-1}, \\
\boldsymbol{\mu}_{w|\mathcal{D}} &= \boldsymbol{\Sigma}_{w|\mathcal{D}} (\boldsymbol{\Sigma}_w^{-1} \boldsymbol{\mu}_w + \mathbf{H}^\top \boldsymbol{\Sigma}_D^{-1} \mathbf{y}_{\mathcal{D}} ).
\end{align}
The posterior mean $\boldsymbol{\mu}_{w|\mathcal{D}}$ defines a trajectory that interpolates the via-points while preserving smoothness.  
We set $\boldsymbol{w}_{\text{ref}} = \boldsymbol{\mu}_{w|\mathcal{D}}$ and let PPO learn residual refinements $\Delta \boldsymbol{w}_t$ on top, effectively separating geometry-constrained adaptation from performance-driven learning.

\subsection{Energy-Tank Layer for Safe Execution}
\label{sec:tank}

To ensure contact safety and passivity, we integrate an \emph{energy-tank mechanism} that monitors instantaneous power 
\[
p_t=\big|{\boldsymbol{\lambda}^{\text{nom}}_t}^\top\boldsymbol{\nu}_t\big|
\]
and scales the nominal command $u_t^{\text{nom}}$ induced by the policy outputs (Sec.~\ref{sec:ppo_promp}) by a factor $\gamma_t \in [0,1]$.
According to the power and energy limits in Eq.~(\ref{eq::problemdefinition}), the executed command
\[
u_t = \gamma_t\,u_t^{\text{nom}}
\]
is constrained by
\[
\gamma_t \le \min\!\left(\frac{P_{\max}}{\max(\varepsilon,p_t)},\ \frac{E_t}{\Delta t\,\max(\varepsilon,p_t)}\right),
\]
with $\gamma_t\in[0,1]$ and $\varepsilon>0$ a small constant.

Thus, the tank state is updated as
\[
E_{t+1} = \min\!\big\{E_{\max},\ \max\{0,\ E_t-\gamma_t p_t \Delta t\}\big\}.
\]
This layer directly scales Cartesian-force commands (extendable to velocity or other channels), ensuring passivity regardless of the policy and stabilizing contact-rich interactions.

\section{Experiments}
\label{sec:experiments}
We design two complementary experiments to validate our method: (i) a pushing task to highlight the smoothness and stability benefits of ProMP-based trajectory generation, and (ii) a 3D maze sliding task to evaluate the generalization capability when facing unseen surface variations.

\subsection{{{Common Experimental Setup}}}
\label{sec:common-setup}

\paragraph{Platform and timing}
All simulations are conducted in the Genesis physics simulator~\cite{Genesis} using a 7-DoF Franka Emika Panda arm. 
Genesis performs dynamics integration at $2$\,kHz, while the high-level controller operates at $100$\,Hz with a fixed timestep of $\Delta t = 0.01$\,s. The same controller frequencies are maintained for real-world experiments to ensure consistency between simulation and hardware.

\paragraph{Policy, actions, and tracking}
We train policies using rsl\_rl PPO with GAE~\cite{rudin2022learning}, employing an actor–critic MLP with ReLU activations (256–256–128) and an adaptive learning rate, over 1000 episodes with parallel environments.  
Two policy parameterizations are considered: (i) \emph{episode-level} ProMP, which outputs trajectory weights, and (ii) \emph{step-wise} PPO, which directly outputs Cartesian velocity commands.  
For the ProMP variants, a Cartesian impedance controller tracks the generated reference trajectories.
Method variants are summarized in Table~\ref{tab:abl_variants}.
\begin{table}[t]
\centering
\begin{tabular}{@{}l l l l@{}}
\toprule
Variant & Policy type & Safety layer & Action \\
\midrule
PP  & Episode-level ProMP & --                      & ProMP parameters \\
\textbf{PPT} & \textbf{Episode-level ProMP} & \textbf{Energy tank \checkmark} & \textbf{ProMP parameters} \\
S  & Step-wise PPO       & --                      & Cartesian velocity \\
ST & Step-wise PPO       & Energy tank \checkmark  & Cartesian velocity \\
\bottomrule
\end{tabular}
\caption{Method variants (PPT is our primary method).}
\label{tab:abl_variants}
\end{table}

\paragraph{Safety constraint (energy tank)}
We enforce a power budget through the energy-tank mechanism described in Sec.~\ref{sec:method}.  
The instantaneous mechanical power is  as $P_t = \boldsymbol{\lambda}_t^\top \boldsymbol{\nu}_t$ and is constrained by $P_t \le P_{\max}$, where $\boldsymbol{\lambda}_t\in\mathbb{R}^6$ is the measured wrist wrench (force/torque) and $\boldsymbol{\nu}_t\in\mathbb{R}^6$ is the end-effector twist. Exceeding this limit incurs a penalty and may result in episode termination.   
The same power-safety mechanism is applied consistently in both simulation and real-world experiments.

\paragraph{Observations and sensing}
Observations consist of joint positions and velocities, end-effector pose and twist, as well as wrist wrench measurements.  
On the hardware, we use the Franka external-wrench estimate, while in simulation we obtain the wrist wrench from Genesis contact reporting at the end-effector link; we denote it by $\boldsymbol{\lambda}_t$ and the corresponding end-effector twist by $\boldsymbol{\nu}_t$.

\paragraph{End-effector tools}

\begin{figure}[h]
    \centering
    \includegraphics[width=0.85\linewidth]{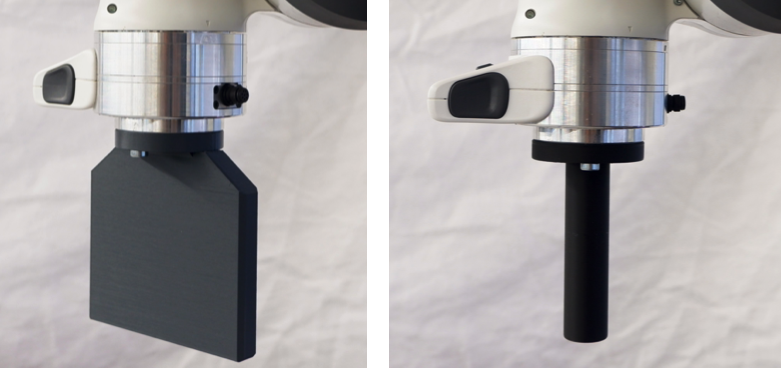}
    \caption{End-effector tools for experiments.}
    \label{fig:tools}
\end{figure}

Two rigid tools are employed (Fig.~\ref{fig:tools}): a slim paddle ($12 \!\times\! 1 \!\times\! 1$\,cm) for the pushing task, and a PLA cylinder (length 12\,cm, diameter 2.5\,cm) for the maze-sliding task.  
The physical maze is fabricated via PLA FDM, intentionally leaving a coarse surface finish to increase contact variability, while the simulation imports the exact STL geometry to match the real-world layout.

\paragraph{Training and reporting protocol}
Unless otherwise specified, results are reported as mean~$\pm$~SE over $k$ random seeds, with curves smoothed using a moving-average window of $w$ steps.  
The episode horizon and success criteria are task-specific and detailed in the corresponding subsections.  
Domain randomization is applied within task-dependent ranges (e.g., friction coefficients, object mass, and  pose) to improve robustness. Shared evaluation metrics are listed in Table~\ref{tab:common-metrics}.

\begin{table}[t]
\centering
\begin{tabularx}{\linewidth}{@{}l l X@{}}
\toprule
Metric & Unit & Definition \\
\midrule
Max Power            & W         & Maximum instantaneous power per episode. \\
Success rate         & \%        & Percentage of episodes satisfying the task success criterion. \\
Jerk RMS             & m/s$^{3}$ & End-effector jerk root-mean-square. \\
Peak wrench (P95)    & N         & $95$th percentile of wrist-wrench norm $\mathrm{P95}(\|\boldsymbol{\lambda}_t\|)$. \\
Overload ratio       & \%        & Time fraction with $P_t > P_{\max}$. \\
Contact continuity   & 0--1      & Mean duration/ratio of contact segments. \\
Progress@T           & 0--1      & Normalized progress at horizon $T$. \\
\bottomrule
\end{tabularx}
\caption{Evaluation metrics shared by all experiments; task-specific thresholds/horizons are defined in each subsection.}
\label{tab:common-metrics}
\end{table}

\subsection{{{Simulation Experiments}}}
\label{sec:sim-exp}

\subsubsection{Box Pushing}
\label{sec:pushing-sim}

A slim paddle is used to push a box across a planar tabletop from a designated start region to a goal.  
Task-specific domain randomization is applied on a per-episode basis: the kinetic friction coefficient is sampled as $\mu_k \sim \mathcal{U}(0.20,0.60)$ (with static friction $\mu_s = 1.25\,\mu_k$), and the box mass is jittered by $\pm 15\%$ around two nominal sizes (6\,cm / 50\,g and 8\,cm / 80\,g).  
During training, the policy has access to privileged box pose information for faster credit assignment; at test time, this information is removed, leaving only start and goal positions (partial observability).
\begin{figure}[b]
    \centering
    \includegraphics[width=\linewidth]{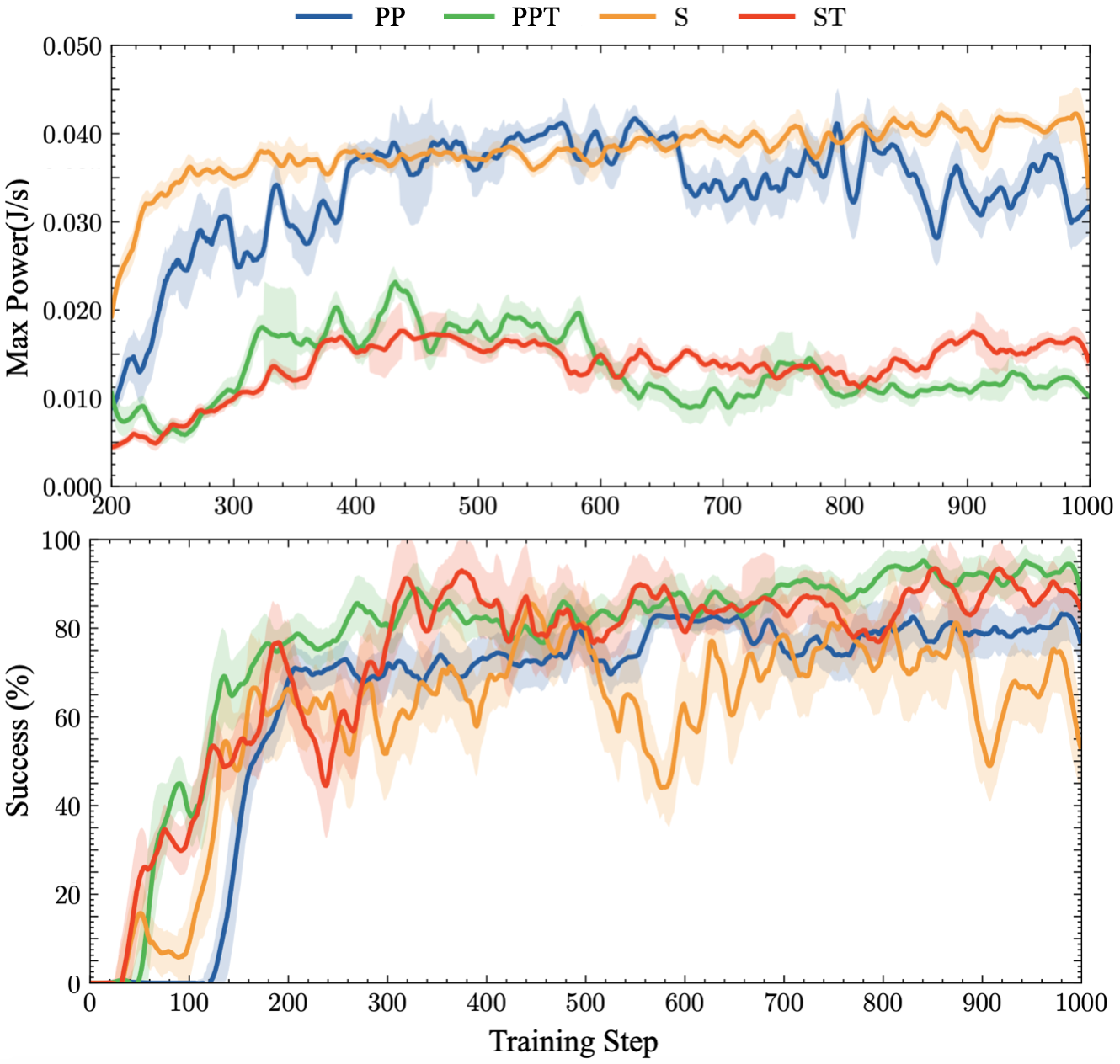}
    \caption{\textbf{Training curves.} Success rate (\%) and max instantaneous power (W) vs.\ training steps for \textbf{PP}, \textbf{PPT}, \textbf{S}, \textbf{ST}. Higher success and lower power are better.}
    \label{fig:pushing_training}
\end{figure}

Across random seeds (Fig.~\ref{fig:pushing_training}), the ProMP-based trajectory policy with energy tank (PPT) exhibits rapid and steady learning, reaching a high success plateau while maintaining the lowest near-peak power. The energy tank effectively clamps force bursts during exploratory contact.  
The step-wise variant with the same safety layer (ST) converges more slowly and shows higher variance due to per-step action fluctuations. A step-wise policy without the tank (PP) learns quickly but suffers occasional regressions, while the step-wise baseline without safety (S) is the least stable.  
Although heavier boxes increase absolute task difficulty, the relative ordering among methods is preserved, demonstrating that trajectory-level generation combined with energy shaping mitigates violent exploration and enhances reliability in contact-rich interactions.

\subsubsection{Maze Sliding}
\label{sec:maze-sim}

\begin{figure}[t]
    \centering
    \includegraphics[width=1\linewidth]{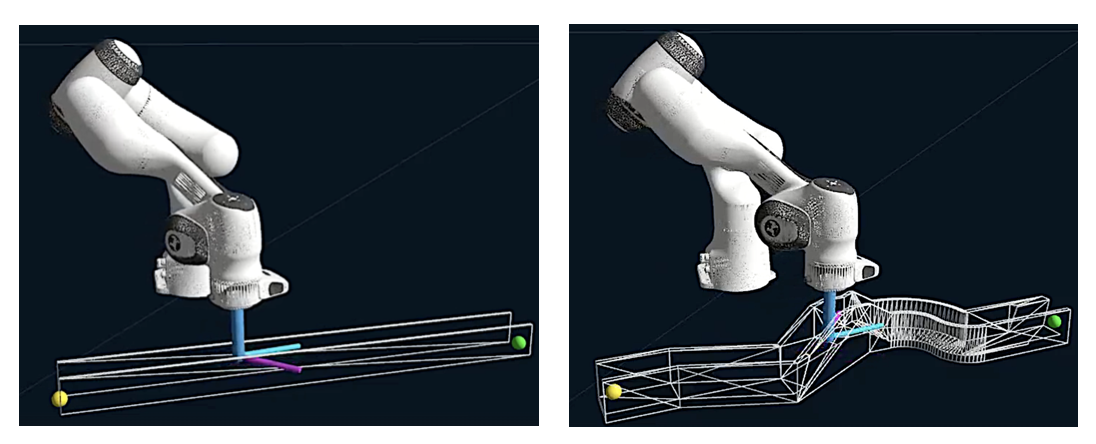}
    \caption{\textbf{Maze sliding scenarios.} Left: training on straight corridors to learn a phase-aligned wall-following prior. Right: deployment in an unseen maze with turns, a disc-like segment, and height variations.}
    \label{fig:maze_train_test}
\end{figure}

Policies are trained exclusively in straight corridors to acquire a phase-aligned wall-following prior, and are subsequently deployed in unseen mazes (corridor width 5–6\,cm, total length $\sim$1\,m) featuring 20°–45° turns, a disc-shaped segment, and up to 4\,cm vertical undulations.  (Fig.~\ref{fig:maze_train_test})  
Observations are limited to proprioception and wrist wrench measurements (no map or vision).  
Domain randomization includes variations in start pose, heading, and bounded wall friction.  
The power budget follows the standard setup, and the episode horizon is $T = 20$\,s.

To isolate the effect of trajectory parameterization, we compare only the deployable, safety-layered variants under the same power budget: PPT and ST.  
Reward components follow the definitions in Sec.~\ref{sec:common-setup}.  
While PPT requires no task-specific redesign, ST achieves consistent performance only after introducing stronger slip and heading regularization.

\begin{figure*}[h]
    \centering
    \includegraphics[width=1\linewidth]{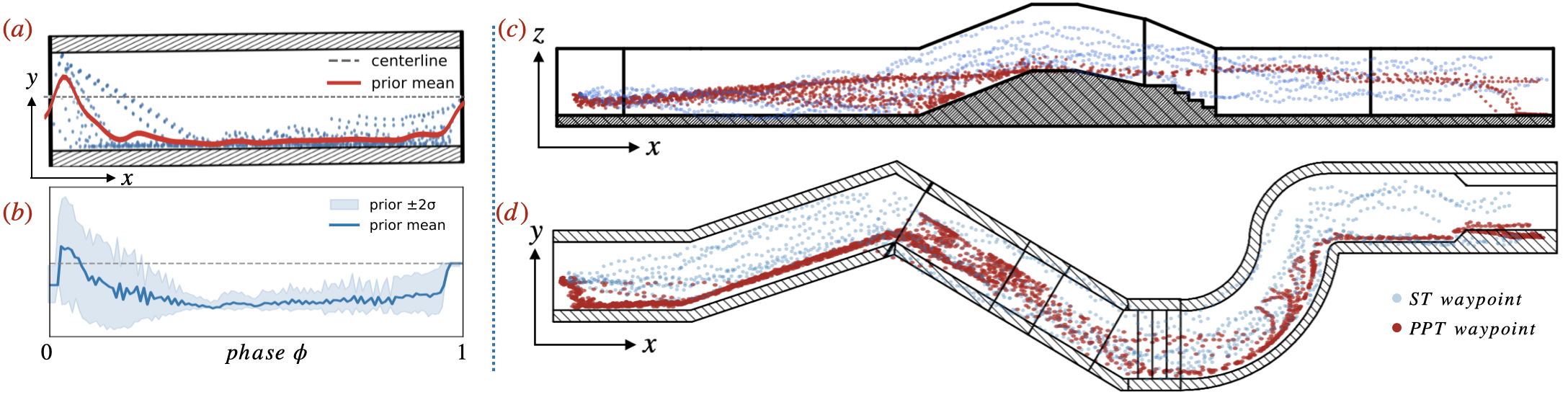}
    \caption{\textbf{Maze sliding—training prior and test posterior rollouts (PPT vs.\ ST).}
    \textbf{(a)} Prior learned on straight corridors: dashed gray centerline and the learned prior mean (solid).
    \textbf{(b)} Prior dispersion over phase $\phi$: mean with $\pm 2\sigma$ band.
    \textbf{(c)} Test maze in $(x\!-\!z)$ view: posterior rollouts with contact-inferred waypoints; \textbf{PPT} in red and \textbf{ST} in blue (dots mark detected waypoints).
    \textbf{(d)} Same as (c) in $(x\!-\!y)$ view. Waypoints cluster near bends/junctions; \textbf{PPT} shows a tighter band and smoother transitions than \textbf{ST}.}
    \label{fig:maze_result}
\end{figure*}

Figure~\ref{fig:maze_result} demonstrates that the wall-following prior learned in straight  successfully transfers to curved and height-varying mazes. Waypoints concentrate near bends, and PPT produces a narrow posterior band that closely adheres to the wall, exhibiting smoother cornering and reduced lateral spread.  

Overall, these results highlight that trajectory-level parameterization combined with energy-aware power gating is crucial for safe, contact-only navigation and effective generalization to novel geometries.

\subsection{{Real-World Experiments}}
\label{sec:real-exp}

\subsubsection{Box Pushing}
\label{sec:real-push}

We evaluate two cubic boxes (edge lengths 6\,cm and 8\,cm; nominal masses 50\,g and 80\,g) on a flat laminate tabletop.  
Surface micro-roughness and dust naturally induce variability in friction and stick–slip behavior, which is \emph{not} explicitly modeled.  
The energy tank remains active with the same budget as in simulation.  
Each trial begins from a randomized start pose within a fixed start region and terminates upon goal capture or safety violation.

\begin{figure}[h]
    \centering
    \includegraphics[width=\linewidth]{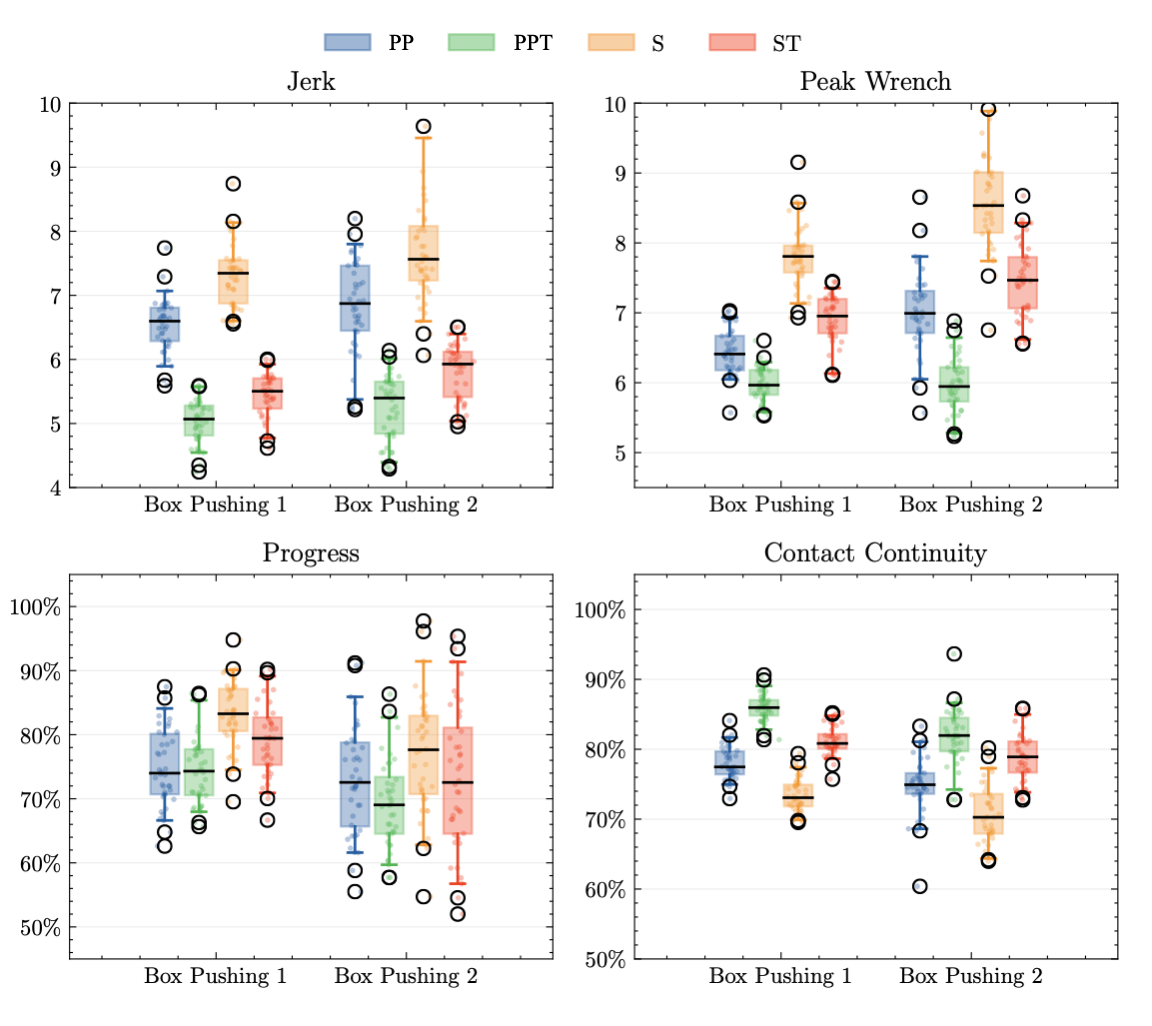}
    \caption{\textbf{Real-world pushing metrics.} 
    Jerk RMS, Peak Wrench P95 (N), Progress (\%), and Contact Continuity (\%) for two object conditions (\emph{Box Pushing 1/2}). 
    Boxes/whiskers show episode/seed distributions.}
    \label{fig:rw-pushing-metrics}
\end{figure}

{PPT} consistently yields the lowest jerk and near-peak wrench and the highest contact continuity, indicating smoother motions and steadier contact under the same power budget (Fig.~\ref{fig:rw-pushing-metrics}). 
\emph{S}/\emph{ST} can reach slightly higher Progress@T on some trials, but at the cost of larger dispersion and more near-overload events, mirroring simulation. 
Absolute jerk/wrench are modestly higher than in sim due to sensing noise and real stick–slip, but the method ranking is preserved, supporting the robustness of trajectory-level parameterization plus power gating.

\subsubsection{Maze Sliding}
\label{sec:real-maze}

We deploy a $1$\,m-long PLA maze with corridor widths of $5$--$6$\,cm, printed with a deliberately coarse surface finish to enhance contact variability.  
The maze layout includes $20^\circ$--$45^\circ$ turns, a disc-like segment, and vertical undulations of up to $4$\,cm.  
The end-effector is a PLA cylinder, and sensing, filtering, and the energy-tank budget are identical to the simulation setup.

\begin{figure}[h]
    \centering
    \includegraphics[width=\linewidth]{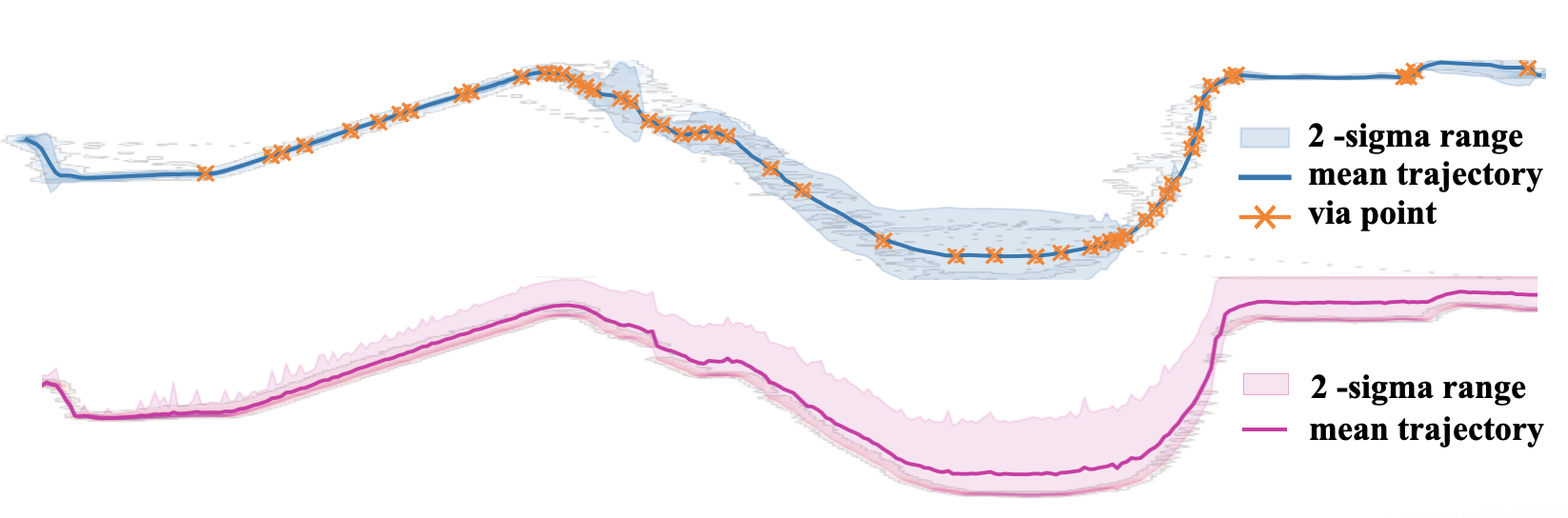}
    \caption{\textbf{Real-world maze sliding.} 
    Top: \textbf{PPT}; bottom: \textbf{ST}. 
    Trajectory mean and $\pm2\sigma$ bands are overlaid; dots denote contact-inferred waypoints. 
    Waypoints cluster near bends/junctions; \textbf{PPT} shows tighter bands and smoother redirection through corners.}
    \label{fig:rw-maze-vis}
\end{figure}

\begin{table}[b]
\centering
\begin{tabular}{@{}l c r r@{}}
\toprule
Metric & Unit & \textbf{PPT} & \textbf{ST} \\
\midrule
Success rate         & \%        & \textbf{89} & 60 \\
Jerk RMS             & m/s$^{3}$ & \textbf{1.85} & 2.70 \\
Peak wrench (P95)    & N         & \textbf{8.5} & 11.2 \\
Overload ratio       & \%        & \textbf{2.3} & 3.1 \\
Contact continuity   & 0--1      & \textbf{0.74} & 0.48 \\
Progress@T           & 0--1      & 0.70 & \textbf{0.76} \\
\bottomrule
\end{tabular}
\caption{Real-world maze sliding under the same power budget: PPT vs.\ ST (means).}
\label{tab:rw-maze-pt-vs-st}
\end{table}

Figure~\ref{fig:rw-maze-vis} and Table~\ref{tab:rw-maze-pt-vs-st} demonstrate that PPT successfully transfers the straight-corridor prior to complex maze geometries, producing trajectories with tighter dispersion and smoother cornering. It achieves higher task success and a stronger safety envelope, reflected in lower jerk, reduced near-peak wrench, fewer overload events, and higher contact continuity. 
ST attains slightly higher progress within a fixed horizon (Progress@T) but exhibits larger lateral spread and more frequent near-overload behaviors. Qualitative observations show that contact-inferred waypoints densify around bends; PPT tracks these transitions with minimal overshoot, whereas ST often oscillates before settling. 
Overall, the sim$\!\rightarrow\!$real trends are consistent: trajectory-level parameterization combined with power gating is key to safe, contact-only navigation, while step-wise control trades stability and safety for speed.

Across both tasks, real-world results corroborate the simulation findings: (i) the energy tank reliably enforces the power budget under unmodeled friction and sensor noise, and (ii) ProMP-based trajectory parameterization yields smoother, more stable behavior with higher success. 
Residual sim-to-real differences, such as slightly higher jerk or wrench, are attributable to measurement noise and surface stick–slip. Notably, PPT required no reward redesign or policy finetuning.

\subsection{{Discussion}}
Our experiments consistently show that the PPT outperforms step-wise policies ST in terms of smoothness and stability. The structured nature of ProMPs promotes globally coherent trajectories, resulting in lower jerk, reduced peak wrench, and higher contact continuity. 
This inherent smoothness mitigates high-frequency, reactive actions that often destabilize step-wise methods during contact.

The results further highlight the synergy between ProMP-based policies and the energy-tank safety layer. 
For PPT, the energy tank acts as a robust safety net for unexpected events, whereas for the more erratic ST policy, frequent tank interventions produce hesitant and inefficient motions, compromising performance despite ensuring safety.

Finally, the framework demonstrates strong practical utility, with robust sim-to-real transfer and generalization to unseen maze geometries without policy finetuning. 
The system effectively handles unmodeled friction and sensor noise, confirming that structured trajectory learning integrated with energy-aware safety is a powerful paradigm for reliable and safe contact-rich manipulation.

\section{Conclusion}

We present a contact-safe reinforcement learning framework that integrates ProMP-based trajectory learning with a passivity-based energy tank. By combining the smooth, structured, and adaptive policies of ProMPs with the strong safety guarantees of the energy tank, our approach enables stable and efficient contact-rich manipulation across a variety of contact surfaces. Numerical simulations and real-world experiments demonstrate that our method outperforms step-wise RL baselines, achieving higher success rates and smoother, safer interactions with robust sim-to-real transfer. Despite these advantages, our method has limitations. The fixed-budget energy tank can be conservative, potentially limiting task performance, and effective generalization relies on a suitable trajectory prior. Future work will explore adaptive energy management strategies to better balance safety and performance, as well as hierarchical trajectory priors to enhance generalization across a broader range of tasks.

\bibliographystyle{IEEEtran}
\bibliography{bibliography}

\end{document}